\documentclass[sigconf,authorversion]{acmart}

\usepackage{booktabs} % For formal tables
\usepackage{graphicx}
\usepackage{amsmath}
\usepackage{booktabs}
\usepackage{float}
\usepackage{multirow}
\usepackage{ulem}
\usepackage{stmaryrd}
\usepackage{siunitx}
\usepackage{textcomp}
\usepackage{titlesec}
\usepackage{xspace}
\usepackage{subfigure}

\usepackage{bm}

\usepackage{color}
\definecolor{url_color}{RGB}{42, 83, 163}
\definecolor{MyPurple}{RGB}{111,0,255}
\definecolor{MyGreen}{rgb}{0.02,0.5,0.02}
\definecolor{MyRed}{rgb}{1.0,0.0,0.0}
\definecolor{MyBlack}{rgb}{0.0,0.0,0.0}
\definecolor{GIF_color}{RGB}{183, 28, 28}

\newcommand{\boldparagraph}[1]{\vspace{0.1em}\noindent{\bf #1} }

\titlespacing{\subsubsection}{0pt}{0.7ex}{0ex}
\titlespacing{\subsection}{0pt}{0.3ex}{0.3ex}
\titlespacing{\section}{0pt}{0.5ex}{0.5ex}

\makeatletter
\DeclareRobustCommand\onedot{\futurelet\@let@token\@onedot}
\def\@onedot{\ifx\@let@token.\else.\null\fi\xspace}

\def\eg{\emph{e.g}\onedot} 
\def\ie{\emph{i.e}\onedot} 
 
\def\etc{\emph{etc}\onedot} 
\def\wrt{w.r.t\onedot} 
\def\etal{\emph{et al}\onedot}
\makeatother

\AtBeginDocument{%
  \providecommand\BibTeX{{%
    \normalfont B\kern-0.5em{\scshape i\kern-0.25em b}\kern-0.8em\TeX}}}

\copyrightyear{2022}
\acmYear{2022}
\setcopyright{acmlicensed}
\acmConference[MM '22] {Proceedings of the 30th ACM International Conference on Multimedia }{October 10--14, 2022}{Lisboa, Portugal.}
\acmBooktitle{Proceedings of the 30th ACM International Conference on Multimedia (MM '22), Oct. 10--14, 2022, Lisboa, Portugal}
\acmPrice{15.00}
\acmISBN{978-1-4503-9203-7/22/10}
\acmDOI{10.1145/3503161.3548125}

\begin{document}

\title{
Factorized and Controllable Neural Re-Rendering of Outdoor Scene for Photo Extrapolation
}

\setlength{\abovedisplayskip}{0.2em}
\setlength{\belowdisplayskip}{0.2em}

\author{Boming Zhao}
\authornote{Boming Zhao and Bangbang Yang contributed equally to this work.
The authors from Zhejiang University are also affiliated with the State Key Lab of CAD\&CG.}
\email{bmzhao@zju.edu.cn}
\affiliation{
  \institution{Zhejiang University}
  \country{}
}

\author{Bangbang Yang}
\authornotemark[1]
\email{ybbbbt@gmail.com}
\affiliation{
  \institution{Zhejiang University}
  \country{}
}

\author{Zhenyang Li}
\email{zhenyounglee@gmail.com}
\affiliation{
  \institution{Baidu.com}
  \country{}
}

\author{Zuoyue Li}
\email{zuli@student.ethz.ch}
\affiliation{
  \institution{ETH Zürich}
  \country{}
}

\author{Guofeng Zhang}
\email{zhangguofeng@zju.edu.cn}
\affiliation{
  \institution{Zhejiang University}
  \country{}
}

\author{Jiashu Zhao}
\email{jzhao@wlu.ca}
\affiliation{
  \institution{Wilfrid Laurier University}
  \country{}
}

\author{Dawei Yin}
\email{yindawei@acm.org}
\affiliation{
  \institution{Baidu.com}
  \country{}
}

\author{Zhaopeng Cui}
\email{zhpcui@gmail.com}
\authornote{
Corresponding authors: Hujun Bao and Zhaopeng Cui.
}
\affiliation{
  \institution{Zhejiang University}
  \country{}
}

\author{Hujun Bao}
\email{bao@cad.zju.edu.cn}
\authornotemark[2]
\affiliation{
  \institution{Zhejiang University}
  \country{}
}

\renewcommand{\shortauthors}{Boming Zhao and Bangbang Yang, et al.}

\begin{abstract}
Expanding an existing tourist photo from a partially captured scene to a full scene is one of the desired experiences for photography applications.
Although photo extrapolation has been well studied, it is much more challenging to extrapolate a photo (\ie, selfie) from a narrow field of view to a wider one while maintaining a similar visual style.
In this paper, we propose a factorized neural re-rendering model to produce photorealistic novel views from cluttered outdoor Internet photo collections, which enables the applications including controllable scene re-rendering, photo extrapolation and even extrapolated 3D photo generation.
Specifically, we first develop a novel factorized re-rendering pipeline to handle the ambiguity in the decomposition of geometry, appearance and illumination.
We also propose a composited training strategy to tackle the unexpected occlusion in Internet images.
Moreover, to enhance photo-realism when extrapolating tourist photographs, we propose a novel realism augmentation process to complement appearance details, which automatically propagates the texture details from a narrow captured photo to the extrapolated neural rendered image.
The experiments and photo editing examples on outdoor scenes demonstrate the superior performance of our proposed method in both photo-realism and downstream applications.
Code and the supplementary material are available on the project webpage:
\urlstyle{tt}
\textcolor{url_color}{\url{https://zju3dv.github.io/neural_outdoor_rerender/}}.
\end{abstract}




\begin{teaserfigure}
  \vspace{-1.7em}
  \includegraphics[width=\textwidth]{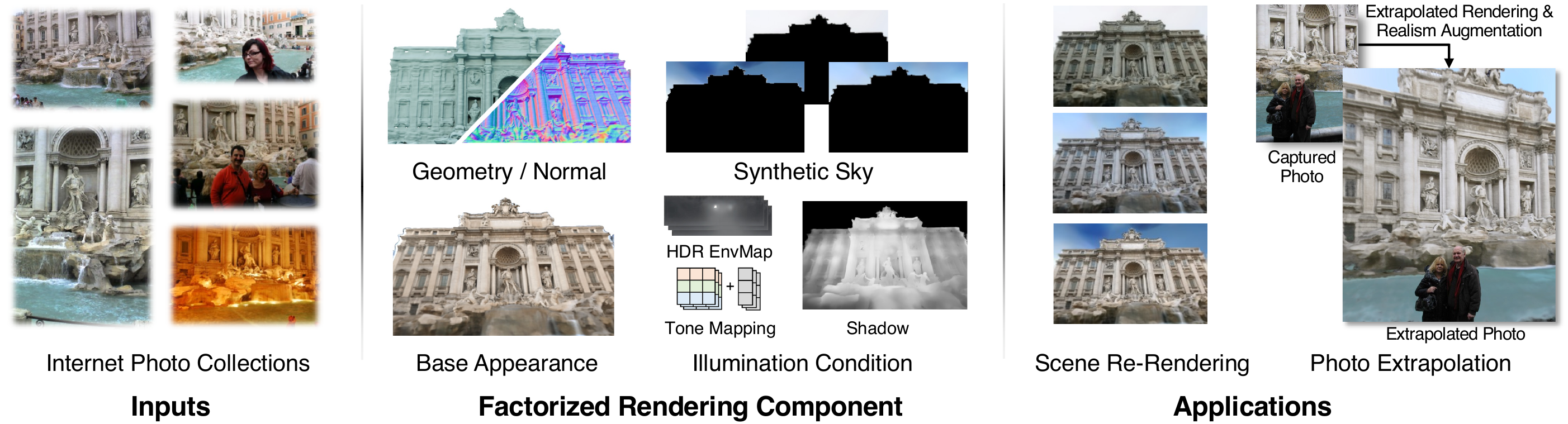}
  \caption{
Given an Internet photo collection of an outdoor attraction, we learn a novel neural re-rendering model that encodes scenes with several factorized components, which enables the applicabilities of controllable scene re-rendering, photo extrapolation and extrapolated 3D photo generation (see GIF in Fig.~\ref{fig:3d_photo}). All the images are from the IMC-PT dataset~\cite{jin2021image}. Photos by Flickr users astrobri, soniadal82, Fotero, scriptingnews, Devin Ford, and MikiAnn. 
  }
%   \Description{Finish it.
%   }
\vspace{-0.0em}
  \label{fig:teaser}
\end{teaserfigure}

\maketitle

\vspace{-1.0em}
\section{Introduction}

When a tourist visits a famous attraction, (s)he usually likes to take a picture that can capture both the person and the whole scene.
However, in many cases, only a part of the scene can be captured due to the narrow field of view (FoV) of the camera and/or crowded people, which may be frustrating to the tourist.
So similar to other image post-processing (\eg, super-resolution, deblurring, \etc), it would be of great help if we can handily extrapolate the photo to obtain a wider view of the scene,
while maintaining a similar visual style (\ie, lighting condition or filtering effect) between the complemented image area and the original photo, as shown in the last row of Fig.~\ref{fig:teaser}.
Moreover, it could be more attractive to expand the extrapolated photo to full-scene dynamic 3D with vivid effects as shown in Fig.~\ref{fig:3d_photo}.

The most straight-forward solution relies on 2D image processing including 2D image stitching~\cite{brown2007automatic},  2D image extrapolation~\cite{biggerpicture,wang2018biggerselfie} or 2D image generation~\cite{pixelsynth,teterwak2019boundless,kim2021painting,sabini2018painting,yang2019very}. 2D image stitching aims to combine multi-view images of small FoVs into a panorama, however it can only generate high-quality images when the input multi-view images are captured at the same location and time, which may be impossible for the crowded environment, selfie mode or the old photos captured before. 
Image extrapolation or generation learns to generate visually consistent content for the extrapolated regions using library images. However, such 2D methods are uncontrollable, which makes the extrapolated parts unreal even though they look plausible, and the library images have to be carefully captured and generally  occlusion-free~\cite{biggerpicture,wang2018biggerselfie}.
Moreover, it is also impossible for these 2D methods to produce 3D photos with immersive user experience (see Fig.~\ref{fig:3d_photo}).

Another possible solution is to first reconstruct the corresponding outdoor scene landmark from a collection of images and then re-render the scene with a large field of view and lighting effect close to the original photos, which has been widely studied in computer vision and graphics~\cite{colmap,XuT19,kazhdan2006poisson,snavely2006photo}.
Traditional approaches use multi-view stereo and photometric blending to reconstruct a textured scene mesh~\cite{kazhdan2006poisson,Waechter2014Texturing}, so as to support the virtual presence of the scene but require high-quality image sets~\cite{Waechter2014Texturing,philip2019multi} and are not feasible for adapting lighting conditions to a specific photograph.
Recently, neural rendering shows promising results on surface mesh reconstructions, novel view synthesis and scene relighting. It also enables to model appearance variations of outdoor scene landmarks from cluttered Internet photo collections~\cite{nrw,nerf_w,block_nerf,ha_nerf}, which liberates the restriction of the data requirement and delivers high flexibility for lighting effect adaptation.
However, existing works either disentangle lighting variations in a latent space while being regardless of explicitly controlling illumination changes~\cite{nerf_w,block_nerf,ha_nerf}, or require carefully captured images to ensure a smooth factorization of material appearances~\cite{nerv,nerd,nerfactor,neural_pil}.
Besides, due to the nature of network smoothness, the rendered images from the neural implicit rendering tend to average the observations and inevitably lose some appearance details (\eg, gushing springs, etc.), which largely degrades the user experience of photo extrapolation.

In this paper, we propose a novel factorized and controllable neural re-rendering pipeline that enables realistic outdoor photo extrapolation from the readily accessible but cluttered Internet photo collections.
Instead of modeling appearance variations as a whole in a latent space~\cite{nrw,nerf_w,block_nerf,ha_nerf}, our rendering model factorizes scene representation into several components (see Fig.~\ref{fig:teaser}), including base appearance, scene geometry (and normal), synthetic sky and an explicitly explainable illumination condition (with data-driven HDR environment map, affine tone mapping and learnable shadow).
Once the scene has been encoded in the rendering model, we can easily perform controllable scene re-rendering under novel views with user-selected lighting conditions, and conduct photo extrapolation or even extrapolated 3D photo generation that extends a captured tourist selfie from a narrow FoV to a widen view, while maintaining similar lighting effect and appearance details by utilizing photo adaptation and a novel realism augmentation mechanism.

However, it is non-trivial to learn a factorized scene representation from cluttered outdoor photos and conduct realistic photo extrapolation with neural rendering models.
\textbf{1)}
Due to the ill-posed nature of the problem, na\"ive solutions of NeRF-based inverse rendering~\cite{nerv,nerd,nerfactor,neural_pil} are no longer applicable for such cluttered images and unbounded outdoor scenes, hence we propose a novel rendering pipeline for this challenging task.
Specifically, at the rendering stage, we utilize a data-driven sky HDR decoder from Gillan \etal~\cite{deepsky} to constrain the HDR map optimization in a reasonable latent space and resolve the scale ambiguity between the recovered HDR environment map and the base appearance.
Then, to model the unobserved shadow caster (\eg, buildings behind the attraction), we introduce a learnable shadow branch that provides spatial shadow value during volume rendering.
Finally, to handle dramatic color distortion that is beyond physically explainable lighting (\eg, user's filtering effect), we apply a learnable affine tone mapping~\cite{urf,block_nerf} to the rendered pixels.
\textbf{2)}
As the training images are collected in a crowd-sourcing paradigm from the Internet, there might be unexpected occluders (\eg, tourists or birds) that affect the learning of factorized rendering even with transient modeling~\cite{nerf_w} (see Sec~\ref{ssec:expr_ablation}).
To tackle this challenge, we employ a composited training scheme to first train the geometry model and then train the rendering model with distilled occlusion-free images from NeRF-W~\cite{nerf_w}.
This process can be regarded as transferring the latent appearance embedding into an explicit and controllable illumination parameterization.
\textbf{3)}
Since the neural implicit model tends to fuse appearances from multi-view observation, the re-rendered scene is somehow more blurry than the user-captured photo and also lacks some live details such as water of the fountain splashes and clouds.
To bridge the gap between the neural rendering and the tourist photos during photo extrapolation, we propose a novel realism augmentation by fully exploiting rich textures from the given photo and propagating them into the rendered view.
In this way, the extrapolated photograph can be more visually coherent to the captured one.

Our contribution can be summarized as follows. First, we propose a novel factorized neural rendering model which learns to encode unbounded outdoor scenes from cluttered Internet photo collections, and delivers the capability of controllable scene re-rendering, photo extrapolation and even extrapolated 3D photo generation. Second, to tackle the challenges of learning outdoor scene representation, our factorized rendering pipeline enables to handle varying lighting effects and color distortions by utilizing a composited training scheme to guide the training process. Moreover, a novel realism augmentation mechanism is also proposed to effectively complement details from a narrow-view real photo to a wide-view synthesized image. At last, the experiments and photography editing examples on several outdoor attractions show the superiority of our method in scene re-rendering, photo extrapolation, and extrapolated 3D photo generation.

\vspace{1.0em}
\section{Related Works}

\noindent\textbf{Outdoor scene reconstruction and rendering.}
Traditional methods generally use SfM~\cite{colmap} and MVS techniques to reconstruct surface mesh~\cite{XuT19,kazhdan2006poisson}, and blend colored images~\cite{Waechter2014Texturing,philip2019multi} to obtain a textured mesh for visualization.
But they require high-quality images with consistent illumination condition, and cannot cannot handle Internet photo collections~\cite{snavely2006photo} with varying lighting and frequent occlusions.
Recently, researchers use neural rendering techniques for outdoor scene rendering~\cite{nerf,li2020crowdsampling,nerf_w,nrw,urf,block_nerf,citynerf,ha_nerf}.
Li \etal~\cite{li2020crowdsampling} uses multi-plane images to render outdoor attractions from photo collections, but cannot produce reasonable views when looking from the tilted side view due to the limitation of MPIs~\cite{nerf}.
NRW~\cite{nrw}, NeRF-W~\cite{nerf_w} and their following works~\cite{urf,block_nerf,citynerf,ha_nerf,yang2022_nr_in_a_room} model outdoor lighting variations with a latent appearance code, which enable novel view synthesis with customizable camera trajectories and support appearance transition with code interpolation.
However, as these methods learn appearance variations in a standalone latent space, they do not support controllable re-rendering with user-selected lighting effects.

\noindent\textbf{Scene rendering with controllable illumination.}
Early methods mainly rely on optical equipment to measure the geometry~\cite{yu1998recovering,loscos1999interactive}, reflectance~\cite{masselus2003relighting,troccoli2008building} and environment lighting~\cite{debevec2006image,stumpfel2006direct} for relightable scene rendering.
In recent years, researchers propose to solve the scene relighting (or inverse rendering) with neural networks~\cite{li2020inverse,li2018cgintrinsics,luo2020niid,yu2020self,yu2021outdoor,zhu2021spatially,deepsky}, but only support static photograph.
Very recently, some works~\cite{nerv,nerd,nerfactor,neroic,neural_pil,guo2020object} build up a relightable implicit representation upon object-centric neural volume rendering~\cite{nerf,yang2021learning,neumesh}, which produce a ``self-occlusion style'' shadow effect by utilizing visibility from learned density field.
However, they cannot be extended to large-scale outdoor scenes and are also not capable of handling noisy observations such as Internet photo collections.
In parallel to our works, Rudnev \etal~\cite{nerf_osr} proposes to learn a neural radiance field for outdoor scene relighting, but shows limited ability of representing and re-rendering on Internet photo collections due to the coarse learned geometry (or surface normal) and simplified lighting model.
Instead, as our surface normal is derived from the SDF field and we utilize a more flexible external lighting model, our approach can be applied to a broader range of outdoor scenes with cluttered photo collections.

\noindent\textbf{Photo extrapolation.}
Photo extrapolation (a.k.a. image outpainting/expansion) can extend a given image with a narrow FoV to a wide view.
Early approaches either build up a photo library~\cite{biggerpicture} or a short video clip~\cite{wang2018biggerselfie} of the surrounding scene, and perform image-montage or stitching to outpaint the images.
Since these approaches use reliable reference images to ensure consistent appearance in the extended area, they require laborious capturing of the scene and cannot adapt to the lighting conditions at different times of the day.
Recent works attempt to conduct the extrapolation with generative neural models~\cite{pixelsynth,teterwak2019boundless,kim2021painting,sabini2018painting,yang2019very}, which shows plausible results for natural landscapes (\eg, mountain valley, beach) or daily scenes (\eg, cars, corridors), but might not look reasonable for tourist attractions with specific shapes and appearances (see Sec.~\ref{ssec:expr_photo_extrapolate}).
Besides, all these image-based photo extrapolation methods are not designed to support rendering novel views with given camera trajectories, which hinders downstream applications like the extrapolated 3D photo generation.

\begin{figure*}[!t]
    \centering
    % \vspace{-1.0em}
    \includegraphics[width=1.0\linewidth, trim={0 0 0 0}, clip]{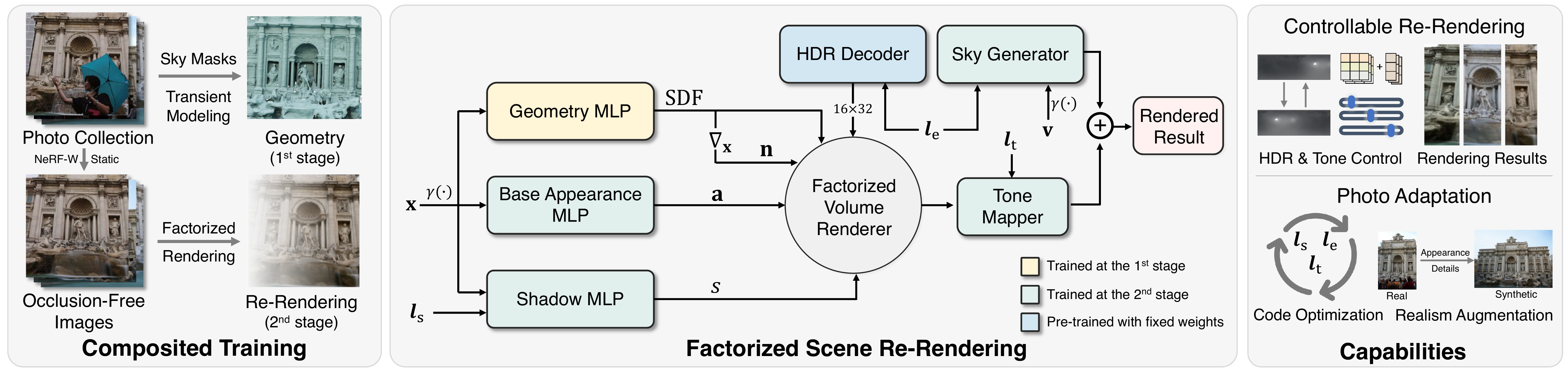}
    \caption{
    Overview.
    Our model learns the geometry and re-rendering of outdoor scenes from the photo collection through a composited training scheme.
    Specifically, scenes are rendered using external lighting with several factorized components, including geometry, basic appearance, HDR environment map, tone mapping, shadows, and synthetic sky.
    See the text for more details.
    Photo by Flickr user chiaki(c\_c).6.}
    \vspace{-1.0em}
    \label{fig:framework}
\end{figure*}

\vspace{1.0em}
\section{Method}

We propose a novel factorized neural rendering framework that learns to encode outdoor scenes from Internet photo collection, which enables controllable scene re-rendering with user-desired lighting condition and photo extrapolation or extrapolated 3D photo generation that extends a narrow-view image to a broaden field of view.
We show an overview of our method in Fig.~\ref{fig:framework}.
Unlike previous neural implicit methods~\cite{nrw,nerf_w,block_nerf,ha_nerf} that encode all the appearance variations (\eg, lighting condition, auto exposure, white balancing and filtering effects, \etc) in one latent space, we %present a first attempting method
present the first attempt
to model outdoor scenes with a more controllable and explainable re-rendering pipeline (Sec.~\ref{ssec:method_render}).
To survive from the training with noisy Internet photos, we utilize a composited training scheme (Sec.~\ref{ssec:method_train}), which learns scene geometry with transient removal strategy from Martin-Brualla \etal~\cite{nerf_w}, and supervises re-rendering with distilled occlusion-free images.
Moreover, we apply a novel realism augmentation technique that propagates appearance details from tourist photos to the rendered views (Sec.~\ref{ssec:method_adaptation}), which efficiently improves the photo-realism of the rendering results.
Please refer to our supplementary material for more technical background.

\subsection{Factorized Outdoor Scene Re-Rendering}
\label{ssec:method_render}

\noindent\textbf{Factorized rendering formulation.}
We first introduce our factorized scene re-rendering pipeline, as shown in the middle part of Fig.~\ref{fig:framework}.
To allow controllable re-rendering of neural implicit model with user-specific external lighting, previous methods generally require empirical normal regularization~\cite{nerfactor,nerv} or post-processing~\cite{neroic} to obtain a smooth surface normal, which inevitably hurts geometry details.
In contrast, we select SDF functions as the representation of scene geometry~\cite{neus}, since it offers exact surface and well-defined normal (by computing gradient \wrt query point) to facilitate the explicit relighting process.
Specifically, we represent scene geometry and basic color with a geometry MLP and base appearance MLP, and use explicit external lighting (\ie, $16\times 32$ HDR map from HDR decoder and affine tone mapping), learnable shadow MLP and a standalone sky generator to re-render the scene with appearance variations.
The rendering of pixel $\hat{C}(\bm{r})$ with point samples $\{\mathbf{x}_k\}$ along the ray $\bm{r}$ is defined as follows:
\begin{equation}
\label{eq:render}
\begin{split}
        \hat{C}(\bm{r}) &= \sum_{k} T_k \alpha_k \Gamma({\mathbf{c}}^b_k, \bm{l}_t) + (1-\sum_{k}  T_k \alpha_k) \mathbf{c}^s(\textbf{v}, \bm{l}_e), \\
        T_k &= \prod_{j=1}^{k-1}(1-{\alpha}_j), \;\;
        {\alpha}_j = \max \left (\frac{\Phi_s(\text{SDF}_{j}) - \Phi_s(\text{SDF}_{j+1})}{\Phi_s(\text{SDF}_j)}, 0 \right),
\end{split}
\end{equation}
where ${\mathbf{c}}^b$ is the relit scene color (introduced later), ${\mathbf{c}}^s$ is the generated sky color along the ray direction $\textbf{v}$ and conditioned by environment code $\bm{l}_e$, $\Gamma(\cdot)$ is the tone mapping conditioned by tone code $\bm{l}_t$, $T$ is accumulated transmittance, $\Phi_s$ is the cumulative distribution of logistic distribution, and $\alpha$ is opacity derived from adjacent SDF.
More specifically, we define the relit scene color ${\mathbf{c}}^b$ as the following:
\begin{equation}
    {\mathbf{c}}^b = \sum _{\omega_k}{\mathbf{a}} s (\bm{l}_s) L_\text{i}(\mathbf{x},\omega_\text{i},\bm{l}_e) (\omega_\text{i} \cdot \mathbf{n}) \Delta{\omega_\text{i}},
\end{equation}
where $\mathbf{a}$ is the basic color from the base appearance MLP,
$s$ is the spatial varying shadow value from the shadow MLP conditioned by shadow code $\bm{l}_s$,
$\omega_\text{i}$ indicate incoming light direction,
$\mathbf{n}$ is the surface normal derived from the gradient \wrt query point $\mathbf{x}$,
$L_\text{i}$ is the incoming HDR lighting along $\omega_\text{i}$ and conditioned by environment code $\bm{l}_e$,
$\Delta{\omega_\text{i}}$ is the solid angle to the light sample.
In summary, the external lighting condition (or appearance variation) is implicitly encoded as environment code $\bm{l}_e$, shadow code $\bm{l}_s$ and tone code.
Note that we adopt the Lambertian reflectance assumption as previous works~\cite{li2020inverse,yu2020self,yu2021outdoor}, which is generally sufficient for outdoor scenes.
Besides, we also apply positional encoding $\gamma(\cdot)$~\cite{nerf} to the query points $\mathbf{x}$ and viewing directions $\mathbf{v}$.
Next, we will introduce the details of each factorized component.

\noindent\textbf{Data-driven HDR decoder.}
Though it is technically sound to re-render outdoor scenes with HDR maps, the problem of disentangling external environment lighting from photo collections is highly ill-posed.
The reasons behind it include the unconstrained freedom of HDR and scale ambiguity between base appearance and HDR intensities.
For example, one might learn an HDR map that embraces main colors of the scene while leaving a degenerated base appearance, or produces a reasonable relit with a brighter base appearance and a darker HDR map.
To this end, we propose to use a data-driven outdoor HDR prior, which constrains the optimization of environment maps in a pre-trained latent space.
Practically, we first train a panoramic HDR sky network~\cite{deepsky} with Laval sky dataset~\cite{deepsky}, and pick up the sky decoder as a prior.
During the re-rendering stage, we fixed the weight of this decoder, and take as input a per-frame latent environment code $\bm{l}_e$, and then downscale the decoder's output to obtain a $16\times 32$ environment map.
In this way, our model can search for a proper HDR map while avoiding color leaking from buildings into light maps (\eg, similar observations for the indoor scenes~\cite{nerfactor}).

\noindent\textbf{Learnable shadow.}
Following standard rendering pipeline~\cite{akenine2019real,sloan2002precomputed}, recent neural implicit rendering and relighting approaches~\cite{nerv,nerfactor,guo2020object} tend to learn a visibility mapping from self-occlusions of density field, which is used to synthesize shadow effect on the rendered views but requires knowing the shape and position of all the object occluders in the scenes.
However, for outdoor photo collections, we found that the shadow might come from unobserved buildings behind the capturing positions, which beyond the visibility from self-occlusion of the reconstructed scene geometry and make it much more complicated.
Rather than pursuing a physically correct shadow mapping, inspired by 2D inverse rendering~\cite{yu2020self}, we propose to model the shadow effect with a spatial variant shadow MLP.
Specifically, the shadow MLP is conditioned with a per-frame latent shadow code $\bm{l}_s$, and learns a shadow value $s\in(0,1)$ for each query point along volume rendering rays.
To alleviate undesired scaling between base appearance and shadow, we add a shadow regularization during the training stage, which encourages the shadow value close to 1 and is defined as the following:
\begin{equation}
\label{eq:rs}
    \mathcal{L}_{\text{rs}} = \sum_k||s_k-1||^2_2.
\end{equation}
In our experiment, we find this shadow modeling successfully simulates shadow effects for cluttered outdoor photo collections while being regardless of issues with unseen shadow casters (Sec.~\ref{ssec:expr_ablation}).

\noindent\textbf{Affine tone mapping.}
Because the output of data-driven HDR decoder is a physically plausible sky lighting that is robust to sensor variation such as white balancing and exposure, we need to take additional tone mapping to handle a large variant of all these color distortions (even including extreme filtering effect by users) in photo collections.
In practice, we adopt the strategy from Rematas \etal~\cite{urf} by learning a $3 \times 4$ affine tone mapping matrix $\begin{bmatrix}A ; \mathbf{b}\end{bmatrix}$  (only upper 3 rows) for each frame, where the matrix is the output of a lightweight tone mapper with per-frame latent tone code $\bm{l}_t$ as input.
To avoid color shifting of base appearance due to unconstrained freedom of mapping matrix, we append an affine regularization into the training loss, which encourages the affine tone mapping to be ``zero-mean'' and is defined as the following:
\begin{equation}
\label{eq:rt}
    \mathcal{L}_{\text{rt}} = \mathbf{e}^\top (R\odot R) \mathbf{e} +  \mathbf{b}^\top \mathbf{b},\; \text{where}\; R=A-I,
\end{equation}
$\odot$ is the Hadamard product, $\mathbf{e}$ is the $3\times1$ column vector whose entries are all 1.

\noindent\textbf{Neural sky generator.}
Since we build up the scene geometry with an SDF-based model, it is not applicable to model background sky along with the buildings~\cite{nerf_w,neus,urf}.
Motivated by Rematas \etal and Hao \etal~\cite{urf,gancraft}, we use a neural sky generator to simulate the sky dome of the scene, which directly maps the viewing direction $\bm{r}$ to a 3-channel sky color with the condition of environment code $\bm{l}_e$.
As demonstrated in Eq.~\eqref{eq:render}, sky colors are blended according to the remaining transmittance, so we can jointly train the sky generator at the re-rendering stage.

\subsection{Composited Training with Photo Collections}
\label{ssec:method_train}
Compared to previous methods, our scene representation is much harder to train due to the ill-posed nature of the factorization and noisy Internet data collections.
Therefore, we develop a composited training, which learns scene geometry and re-rendering in a two-staged fashion.

\noindent\textbf{Learning scene geometry from Internet photos.}
In the first stage, we learn scene geometry from photo collections with a geometry MLP and radiance MLP following Wang \etal~\cite{neus}.
To handle occasional object occlusions and appearance variations, we adopt the appearance embedding and transient MLP from NeRF-W~\cite{nerf_w}  (only for this stage).
Unlike radiance field methods~\cite{nerf_w,block_nerf} that can render the sky with scattered far sampling, the SDF-based method is inclined to learn an exact surface, which results in a sky dome stitching to the building that is not desired for external relighting.
So, we additionally apply a sky segmentation loss to encourage the sky area to be empty, which is defined as:
\begin{equation}
    \mathcal{L}_{\text{m}} = \text{BCELoss}\left(\sum_{i}  T_k \alpha_k , M_{\text{sky}}\right),
\end{equation}
where the sky mask $M_{\text{sky}}$ is annotated with Mask-RCNN~\cite{he2017mask}.
Now, we define the training loss of the scene geometry as:
\begin{equation}
    \mathcal{L}_{\text{geo}} = 
    \lambda_{\text{c}}\mathcal{L}_{\text{c}} +
    \lambda_{\text{m}}\mathcal{L}_{\text{m}} +
    \lambda_{\text{re}} \mathcal{L}_{\text{re}},
\end{equation}
where $\mathcal{L}_{\text{c}}$ is the photometric loss with transient modeling following NeRF-W~\cite{nerf_w},  $\mathcal{L}_{\text{re}}=||\left\lVert\nabla_{\textbf{x}} \text{SDF}(\textbf{x})\right\rVert-1||^2_2$ is the Eikonal loss as suggested by Gropp \etal~\cite{gropp2020implicit}.
We set $\lambda_{\text{c}}=1.0$, $\lambda_{\text{m}}=0.1$ and $\lambda_{\text{re}}=0.1$.
Note that we omit the form of pixels summation in this section for brevity.
After the first training stage, we only keep the geometry MLP, while discarding radiance and transient MLP.

\noindent\textbf{Joint optimization of re-rendering.}
In the second stage, we learn factorized scene re-rendering with geometry MLP frozen.
Instead of training with raw cluttered photos or manually masking out occluders, we take a distilled fashion by exploiting occlusion-free images from NeRF-W's static branch for a more steady supervision.
Our experiment shows that this strategy efficiently eases the learning of factorized re-rendering and improves the rendering quality both quantitatively and qualitatively (Sec.~\ref{ssec:expr_ablation}).
The training loss is then defined as the following:
\begin{equation}
    \mathcal{L}_{\text{render}} = 
    \lambda_{\text{cr}} \mathcal{L}_{\text{cr}} +
    \lambda_{\text{rs}} \mathcal{L}_{\text{rs}} +
    \lambda_{\text{rt}} \mathcal{L}_{\text{rt}} ,
\end{equation}
where $\mathcal{L}_{\text{cr}}$ is the MSE loss between the re-rendered pixel and the occlusion-free images.
$\lambda_{\text{cr}}, \lambda_{\text{rs}}$, and $\lambda_{\text{rt}}$ are the loss weights for the MSE loss, shadow regularization (see Eq.~\eqref{eq:rs}), and affine tone mapping regularization (see Eq.~\eqref{eq:rt}) respectively.
We empirically set $\lambda_{\text{cr}}=2.0$, $\lambda_{\text{rs}}=0.01$ and $\lambda_{\text{rt}}=0.1$.

\begin{figure}[!t]
    \centering
    \includegraphics[width=1.0\linewidth, trim={0 0 0 0}, clip]{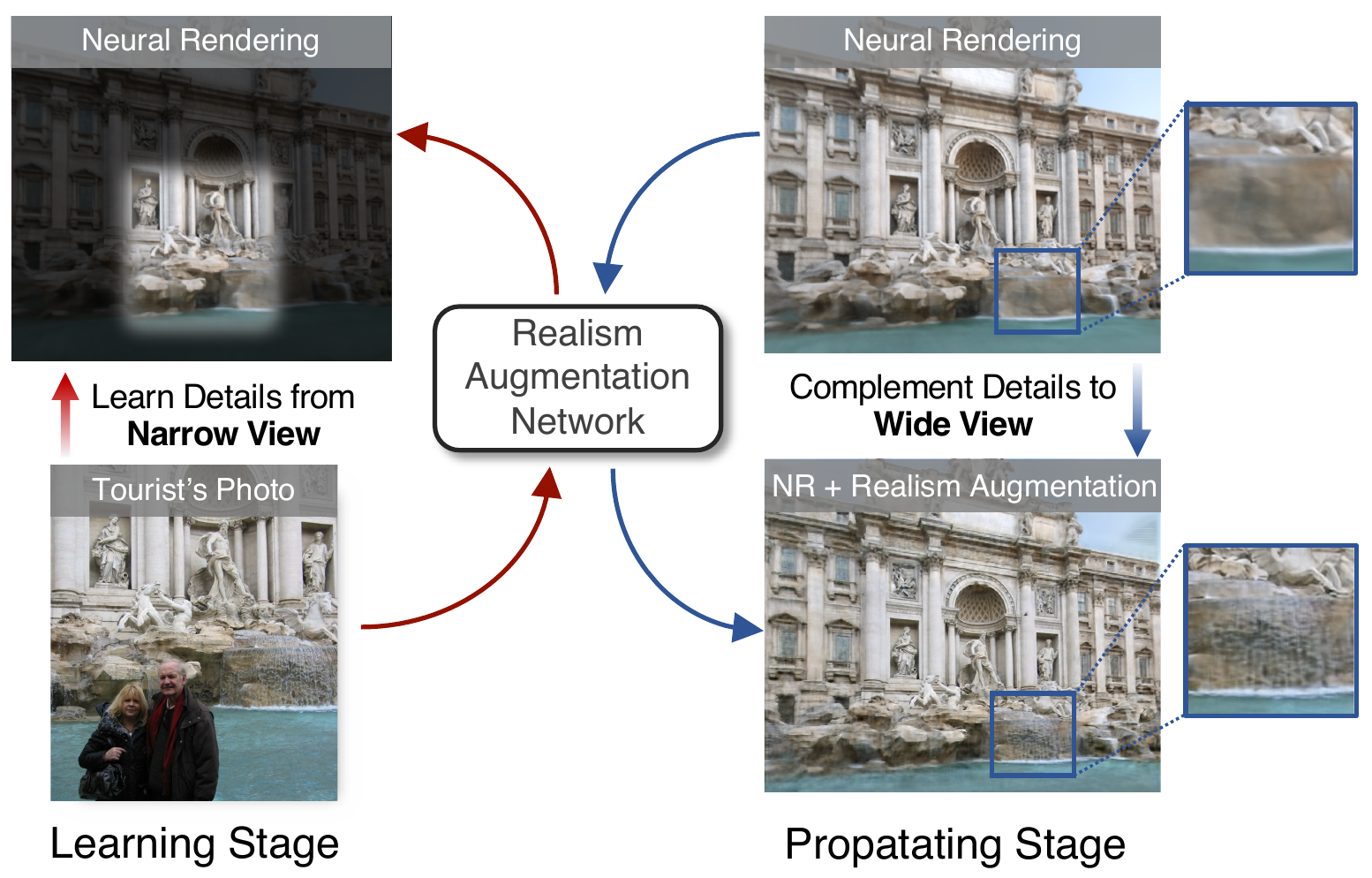}
    \caption{The pipeline of realism augmentation.
    We exploit texture details from tourist's photos and propagate these details (\eg, water splash of the fountain) into the neural rendered large-FoV image. Photo by Flickr user MikiAnn.
    }
    \vspace{-1em}
    \label{fig:pipeline_adaptation}
\end{figure}

\subsection{Photo Adaptation \& Realism Augmentation}
\label{ssec:method_adaptation}

\noindent\textbf{Optimization based photo adaptation.}
Once the factorized scene representation has been trained, our model can be adapted to real-captured photos with novel lighting conditions, \ie, minimizing photometric error between rendered pixels and the captured photo with latent optimization on shadow code $\bm{l}_s$, environment code $\bm{l}_e$ and tone code $\bm{l}_t$.
Besides, when adapting to photos with a large portion of people like tourist selfies in photo extrapolation applications, we empirically mask out these part and only optimize pixels labelled as sky and attraction (\eg, buildings and sculptures).

\noindent\textbf{Realism augmentation.}
Even though the optimization-based photo adaptation can achieve a rendering result close to the real photo, the detail of the synthetic view is still less realistic.
For example, some live details such as water splashes of the fountain and shapes of clouds are missing, which is mainly due to the fact that the neural implicit rendering tends to average texture details from multi-view observations. 
Fortunately, for tasks like photo extrapolation, there is still an opportunity to enhance rendering details if we can fully exploit information from the given photos.
To this end, we design a novel realism augmentation strategy, which significantly propagates texture details from a narrow-view real photo to a wide-view rendered image.
We show the pipeline of this strategy in Fig.~\ref{fig:pipeline_adaptation}, which illustrates the on-the-fly learning and inference fashion of this augmentation process.
Specifically, we adopt an encoder-decoder based network structure from a super-resolution work LIIF~\cite{chen2021learning} as the realism augmentation network since it uses an implicit representation of the image and is flexible to support arbitrary scale and aspect ratio.
At the learning stage, we set the network input as a downscale rendered image, and fine-tune the network with the target of aligned and masked (without occluders such as tourists) real photo, so the network learns to compensate details from a blurry neural rendering to the real one.
Then, at the propagating stage, we fix the network and simply forward the network with the complete rendered image, where the learned ``detail-compensating'' knowledge would be propagated to the full view of the rendering result.

\begin{figure*}[!t]
    \centering
    \includegraphics[width=1.0\linewidth, trim={0 0 0 0}, clip]{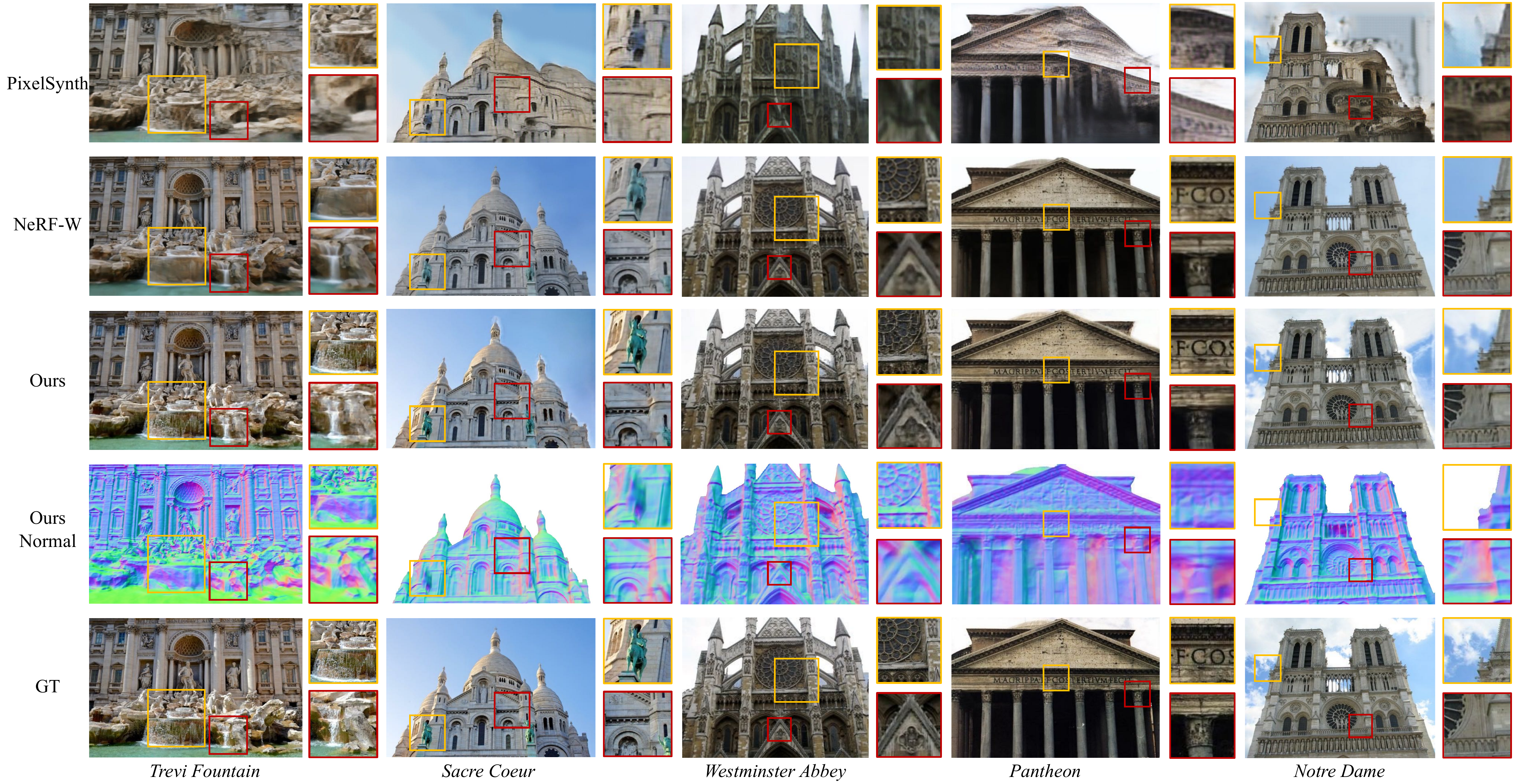}
    \caption{We compare the scene re-rendering quality with other methods on several outdoor attractions, and also visualize the surface normal of our modeling.
    Note that our normal is much smoother than the previous NeRF-based method (see the supplementary materials for the comparison with NeRF-W). All images are from~\cite{jin2021image}.
    }
    \vspace{-0.5em}
    \label{fig:rendering_quality}
\end{figure*}

\begin{table*}[t!]
\centering
\resizebox{1.0\linewidth}{!}{
\begin{tabular}{lccccccccccccccc}
\toprule
\multicolumn{1}{c}{\multirow{2}{*}{Methods}} & \multicolumn{3}{c}{\textit{Trevi Fountain}} & \multicolumn{3}{c}{\textit{Sacre Coeur}} & \multicolumn{3}{c}{\textit{Pantheon Exterior}} & \multicolumn{3}{c}{\textit{Westminster Abbey}} & \multicolumn{3}{c}{\textit{Notre Dame Front Facade}}
\\ \cmidrule(lr){2-4} \cmidrule(lr){5-7} \cmidrule(lr){8-10} \cmidrule(lr){11-13} \cmidrule(lr){14-16} 
\multicolumn{1}{c}{} & \multicolumn{1}{l}{PSNR $\uparrow$} & \multicolumn{1}{l}{SSIM $\uparrow$} & \multicolumn{1}{l}{LPIPS $\downarrow$} & \multicolumn{1}{l}{PSNR $\uparrow$} & \multicolumn{1}{l}{SSIM $\uparrow$} & \multicolumn{1}{l}{LPIPS $\downarrow$} & \multicolumn{1}{l}{PSNR $\uparrow$} & \multicolumn{1}{l}{SSIM $\uparrow$} & \multicolumn{1}{l}{LPIPS $\downarrow$} & \multicolumn{1}{l}{PSNR $\uparrow$} & \multicolumn{1}{l}{SSIM $\uparrow$} & \multicolumn{1}{l}{LPIPS $\downarrow$} & \multicolumn{1}{l}{PSNR $\uparrow$} & \multicolumn{1}{l}{SSIM $\uparrow$} & \multicolumn{1}{l}{LPIPS $\downarrow$} 
\\ \hline
PixelSynth~\cite{pixelsynth} & 14.73 & 0.587 & 0.693 & 14.19 & 0.659 & 0.566 & 12.09 & 0.603 & 0.607 & 14.00 & 0.664 & 0.618 & 12.43 & 0.589 & 0.643 \\
% NeRF-W *~\cite{nerf_w} 
NeRF-W *~\cite{nerf_w}  & 21.31 & 0.764 & 0.380 & 21.23 & \textbf{0.850} & 0.283 & \textbf{24.78} & \textbf{0.875} & 0.265 & 21.35 & 0.801 & 0.364 & 20.86 & 0.735 & 0.456 \\
Ours & \textbf{23.09} & \textbf{0.792} & \textbf{0.345} & \textbf{21.51} & 0.849 & \textbf{0.162} & 24.46 & 0.867 & \textbf{0.237} & \textbf{24.66} & \textbf{0.854} & \textbf{0.238} & \textbf{22.75} & \textbf{0.833} & \textbf{0.254} \\
\bottomrule
\end{tabular}
}
\caption{
We compare the scene re-rendering quality with PixelSynth~\cite{pixelsynth} and NeRF-W~\cite{nerf_w} on five outdoor scenes of the Internet photo collections~\cite{jin2021image,snavely2006photo}.
Note that we use an alternative implementation of NeRF-W. See the text for details.
}
\label{tab:render}
\vspace{-1.5em}
\end{table*}

\section{Experiments}

In this section, we first evaluate the outdoor scene re-rendering quality of our method (Sec.~\ref{ssec:expr_rerender}), and then conduct photo extrapolation (Sec.~\ref{ssec:expr_photo_extrapolate}), controllable scene re-rendering (Sec.~\ref{ssec:expr_control_rerender}) and extrapolated 3D photo generation (Sec.~\ref{ssec:expr_3d_photo}) on several outdoor attractions.
At last, we perform ablation studies to analyse the effectiveness of the training strategy and factorized re-rendering components (Sec.~\ref{ssec:expr_ablation}).

\subsection{Datasets}

Following the previous work~\cite{nerf_w}, we use Internet photo collections of outdoor attractions from the Phototourism (IMC-PT) 2020 dataset~\cite{jin2021image,snavely2006photo}, where the image poses are recovered by COLMAP~\cite{colmap}.
Specifically, we select 5 famous tourist attractions, including \textit{Trevi Fountain}, \textit{Sacre Coeur}, \textit{Westminster Abbey}, \textit{Pantheon} and \textit{Notre Dame}.
For \textit{Trevi Fountain} and \textit{Sacre Coeur}, we follow the split of NeRF-W for training and testing.
For the other three scenes, we also take a similar pre-processing by discarding training images with a large portion of object occlusions, and only select occlusion-free images for metric evaluation.

\begin{figure*}[!t]
    \centering
    \includegraphics[width=1.0\linewidth, trim={0 0 0 0}, clip]{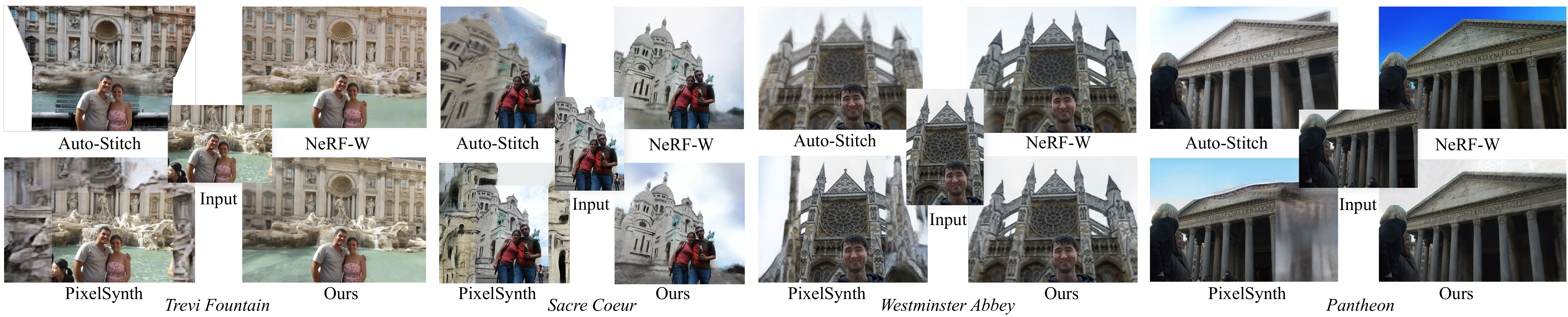}
    \vspace{-1.5em}
    \caption{We compare photo extrapolation with Auto-Stitch~\cite{brown2007automatic}, PixelSynth~\cite{brown2007automatic} and NeRF-W~\cite{brown2007automatic} on four outdoor scenes~\cite{jin2021image,snavely2006photo}. Photos by Flickr users Hugão Cota, Legalv1, Foster's Lightroom, and stobor.
    }
    \label{fig:rendering_photo_extra}
    \vspace{-1em}
\end{figure*}

\subsection{Comparison of Scene Re-Rendering Quality}
\label{ssec:expr_rerender}

We first compare the scene re-rendering quality with the evaluation protocol from NeRF-W~\cite{nerf_w}, \ie, giving a left half image for optimization, the neural network is asked to render the full view of the image. The metrics of PSNR, SSIM and LPIPS~\cite{zhang2018perceptual} are used to measure the rendering quality.
Specifically, we adopt the baseline method NeRF-W~\cite{nerf_w} and the SOTA image extrapolation method PixelSynth~\cite{pixelsynth} for comparison.
The other relightable neural implicit rendering methods (\eg, NeRV~\cite{nerv}, NeRD~\cite{nerd}, NerFactor~\cite{nerfactor}, and \etc) are not applicable here, because they are not feasible for unbounded outdoor scenes or learning from cluttered photo collections.
Note that since NeRF-W has not released the official source code, we adopt an alternative implementation \footnote{\url{https://github.com/kwea123/nerf_pl/tree/nerfw}} in our experiment, thus the reported result is different from~\cite{nerf_w}.
For PixelSynth, we overfit the network to each individual scene with occlusion-free training images as introduced in Sec.~\ref{ssec:method_train}.
We report the quantitative results in Tab.~\ref{tab:render} and present the quantitative visualization in Fig.~\ref{fig:rendering_quality}.
It is obvious that even though we overfit each PixelSynth model to a specific scene, the complete rendering view is still far from satisfactory (\eg, the geometry structure differs a lot to the ground-truth), which proves that the GAN network of the PixelSynth does not ensure a consistent rendering output.
NeRF-W achieves much better results for scene re-rendering, but some live details (\eg, water splash in \textit{Trevi Fountain} and the sky clouds in \textit{Notre Dame}) are still missing due to the smooth nature of neural implicit field~\cite{nerfactor}.
Generally, a disentangled rendering pipeline is usually more challenging to render high-quality images~\cite{nerfactor,nerv}.
Thanks to the factorized re-rendering pipeline and realism enhancement in our method, we still achieve on-par or even better rendering quality both quantitatively and qualitatively while successfully maintaining live details close to the ground-truth.
Moreover, we also exhibit our surface normal in the fourth row of Fig.~\ref{fig:rendering_quality}, which demonstrates the high-quality geometry of the learned model and we believe it is the key to achieving a good re-rendering result with external lighting.
Please refer to the supplementary materials for the additional comparison of surface normal with NeRF-W.

\begin{figure}[!t]
    \centering
    \includegraphics[width=1.0\linewidth, trim={0 0 0 0}, clip]{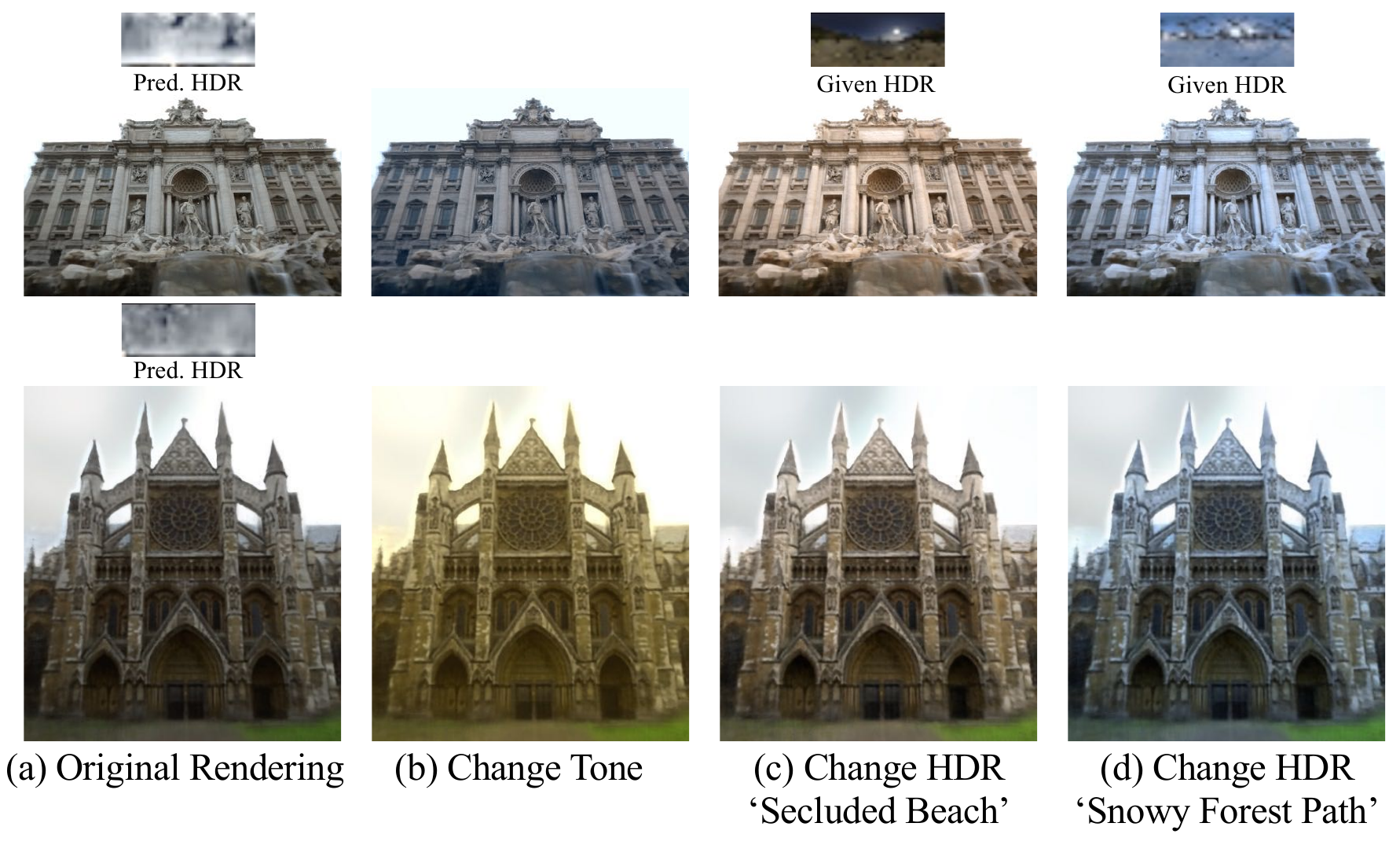}
    \vspace{-1.5em}
    \caption{We exhibit the results of controllable re-rendering by changing tone mapping and HDR environment maps.
    }
    \vspace{-0.75em}
    \label{fig:controllable_rerender}
\end{figure}

\subsection{Comparison of Photo Extrapolation}
\label{ssec:expr_photo_extrapolate}

We now conduct the comparison on the photo extrapolation task in Fig.~\ref{fig:rendering_photo_extra}.
In this part, we also perform an extrapolation test on a 2D image-based method~\cite{brown2007automatic}, which is denoted as Auto-Stitch.
Specifically, we first retrieve $\sim$30 nearby captured position from photo collections based on the SfM camera poses and perform panoramic stitching with OpenPano~\footnote{\url{https://github.com/ppwwyyxx/OpenPano}}~\cite{brown2007automatic}.
Then, we warp the stitched image to the given photo view and blend the front tourists into the image.
It is clear that the extrapolated views of this pipeline are full of stitching artifacts such as human shadows, and there are some vacancy parts near the border of extended images due to the lack of observations near the captured position, which indicates that the 2D image-based approach is not suitable for photo extrapolation with Internet photo collections, as they might need a carefully collected data library to achieve a clean result~\cite{biggerpicture,philip2019multi}.
For the photo extrapolation results of NeRF-W and PixelSynth, similar to what we have analyzed in Sec.~\ref{ssec:expr_rerender}, they are faced with the issues of the lack of live texture details and distorted 3D structure and color (\eg, in \textit{Pantheon} of Fig.~\ref{fig:rendering_photo_extra}, due to the generalizability issue of appearance embedding, NeRF-W correctly simulates the appearance of the building, but the blue sky color is severely distorted), which inevitably degrades the user experience of this functionality.
Thanks to the factorized scene representation and realism augmentation, our results show better photo-realism with vivid details.

\begin{figure}[t!]
\vspace{-0.5em}
\centering
\subfigure[Original Static Photo]{
\begin{minipage}[t]{0.37\linewidth}
\centering
\includegraphics[width=1.0\linewidth, trim={0 0 0 0}, clip]{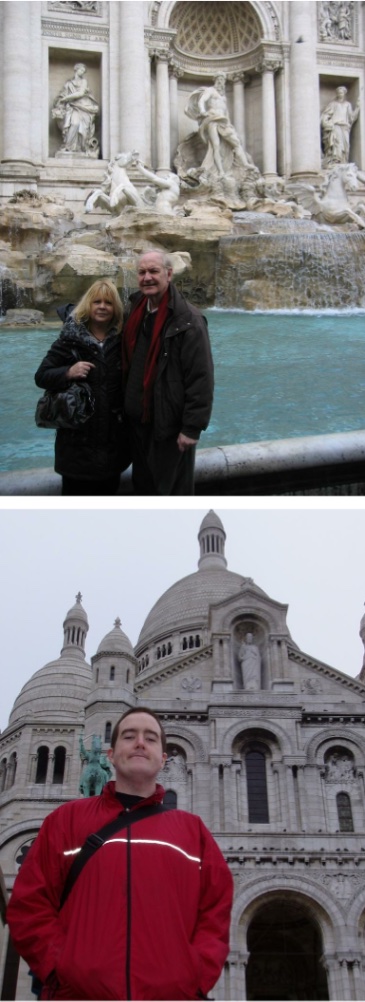}
\end{minipage}
}%
\subfigure[\textcolor{GIF_color}{\textbf{[GIF] }}Extrapolated 3D Photo]{
\begin{minipage}[t]{0.613\linewidth}
\centering
% \animategraphics[width=1.0\linewidth, autoplay, loop]{10}{figures/3d_photo/merge/pic}{1}{30}
\includegraphics[width=1.0\linewidth, trim={0 0 0 0}, clip]{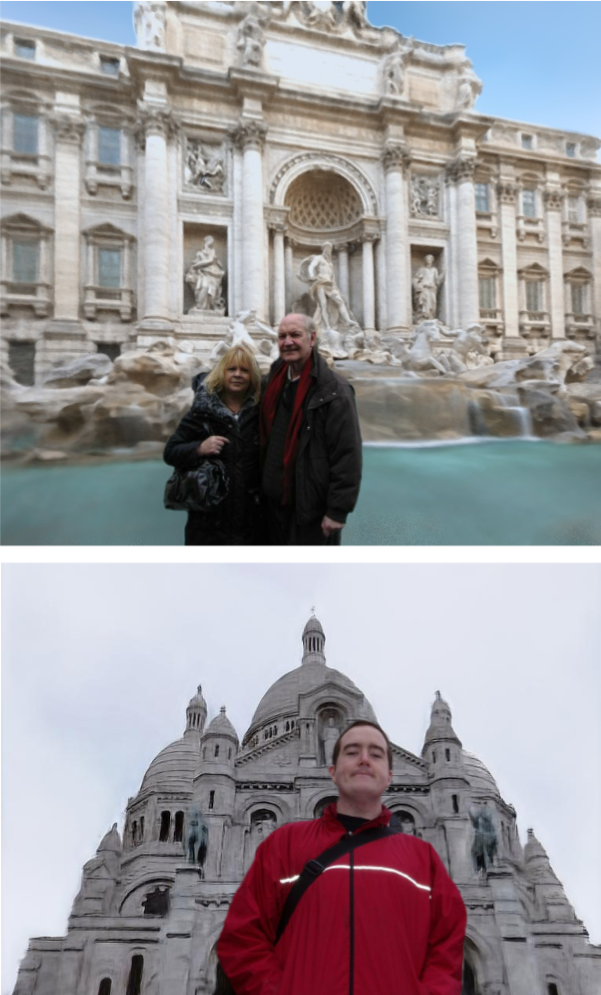}
\end{minipage}
}%
\centering
\vspace{-0.75em}
\caption{We show two examples of extrapolated 3D photo generation, which transfers tourist photos into extrapolated and dynamic 3D photos with camera moving effect. 
Please use \textcolor{GIF_color}{Adobe Reader} or check our \href{https://zju3dv.github.io/neural_outdoor_rerender/}{\textcolor{GIF_color}{project webpage}} to see animations.
Photos by Flickr users MikiAnn and Chris Devers.}
\vspace{-1.5em}
\label{fig:3d_photo}
\end{figure}

\subsection{Controllable Scene Re-Rendering}
\label{ssec:expr_control_rerender}

We show our controllable scene re-rendering capability with user-selected tone mapping and HDR environment maps in Fig.~\ref{fig:controllable_rerender}.
Note that we cannot find proper competitors for this task,
since existing methods either only support appearance changing through the latent space while lack of explicit controlling of lighting effect~\cite{nrw, nerf_w,block_nerf,ha_nerf}, or only support inverse rendering and relighting for 2D images but cannot synthesize novel lighting effect with given camera trajectories~\cite{yu2021outdoor,yu2020self}. 
As shown in Fig.~\ref{fig:controllable_rerender}, with the factorized scene representation, we can freely control the lighting effect through tone mapping and even user-selected HDR maps, \eg, the appearance of the re-rendered building naturally exhibits the lighting effect with a dusk and cool tone from the given HDR maps in Fig.~\ref{fig:controllable_rerender} (c) and (d).

\subsection{Extrapolated 3D Photo Generation}
\label{ssec:expr_3d_photo}
We show the capability of extrapolated 3D photo generation in Fig.~\ref{fig:3d_photo}.
As shown in Fig.~\ref{fig:3d_photo}, by simply adapting lighting condition to the given photo (Sec.~\ref{ssec:method_adaptation}) and enlarging FoV of the renderer, our method naturally transfers a static tourist photo into an extrapolated and dynamic 3D photo with vivid camera moving effect, whereas previous works~\cite{Shih3DP20,wiles2020synsin,niklaus20193d} only generate 3D photo bounded by visible areas.
Please refer to the supplementary material for the detailed implementation of our 3D photo generation.

\subsection{Ablation Studies}
\label{ssec:expr_ablation}

\begin{figure}[!t]
    \vspace{-1.5em}
    \centering
    \includegraphics[width=1.0\linewidth, trim={0 0 0 0}, clip]{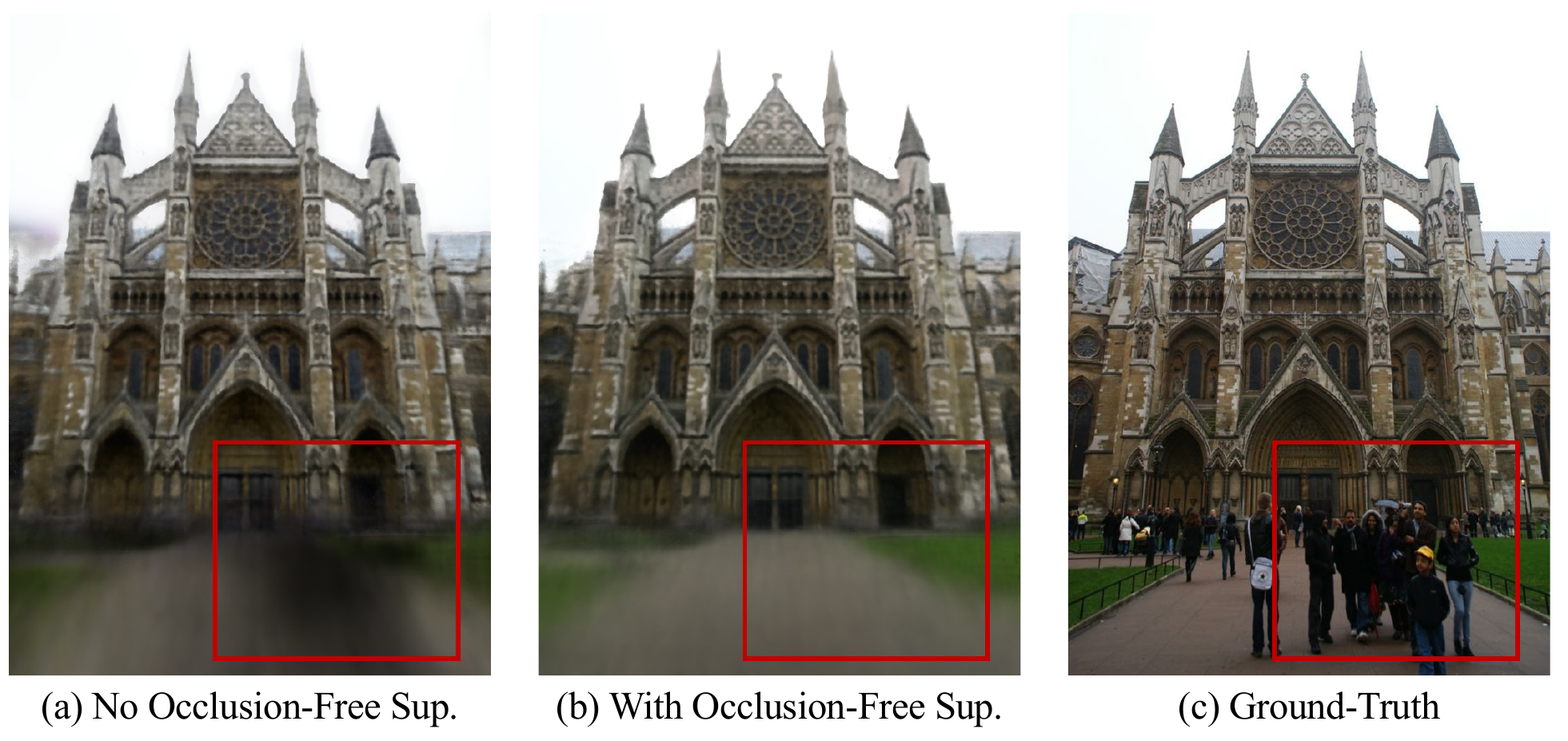}
    \vspace{-1.5em}
    \caption{
    We inspect the trained model w/ or w/o occlusion-free supervision for the re-rendering training stage.
    }
    \vspace{-1em}
    \label{fig:ablation_occlusion_free}
\end{figure}

\begin{figure}[!t]
    \centering
    \includegraphics[width=1.0\linewidth, trim={0 0 0 0}, clip]{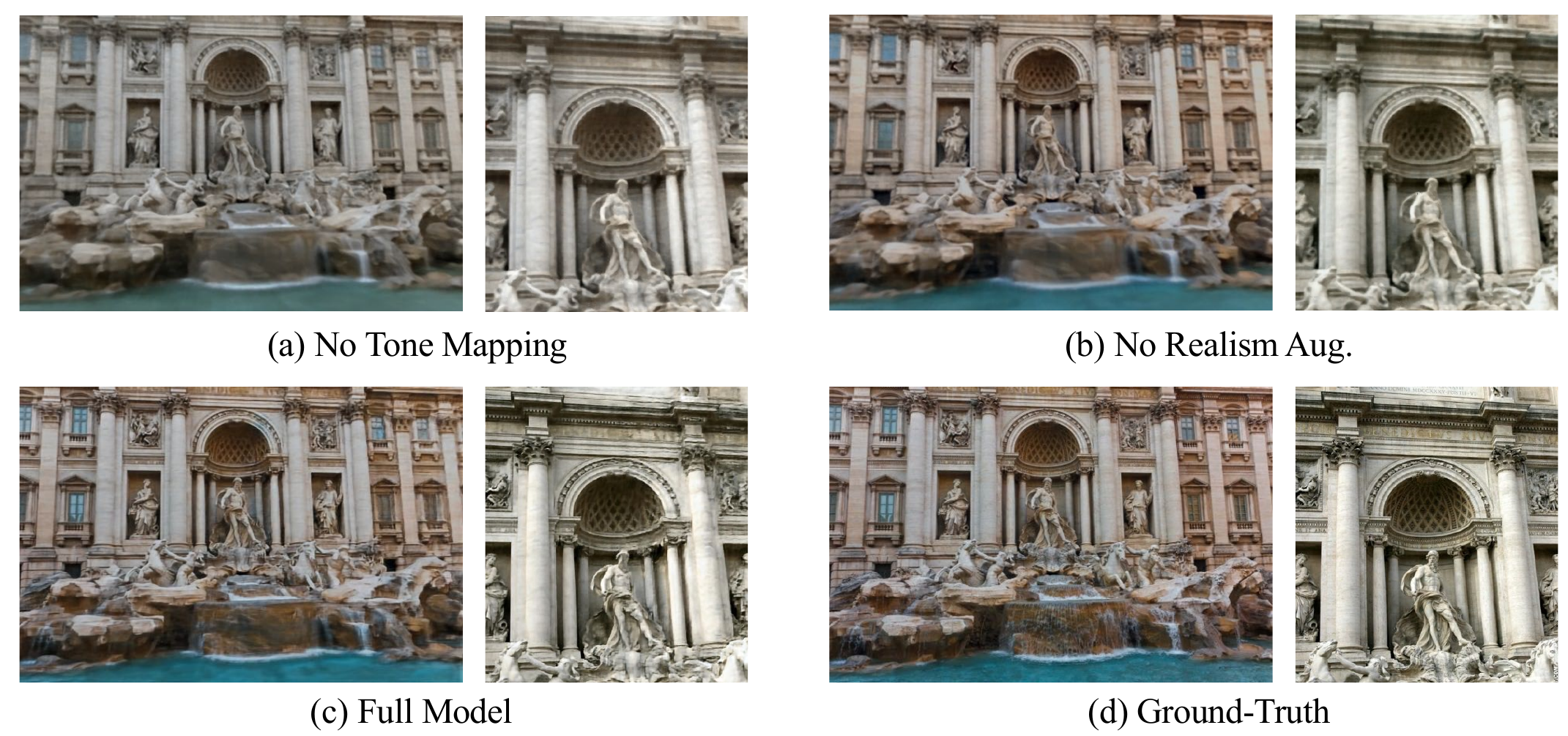}
    \vspace{-1.5em}
    \caption{
    We analyse the effectiveness of affine tone mapping and realism augmentation.
    }
    \vspace{-1em}
    \label{fig:ablation}
\end{figure}

\begin{figure}[!t]
    \centering
    \includegraphics[width=1.0\linewidth, trim={0 0 0 0}, clip]{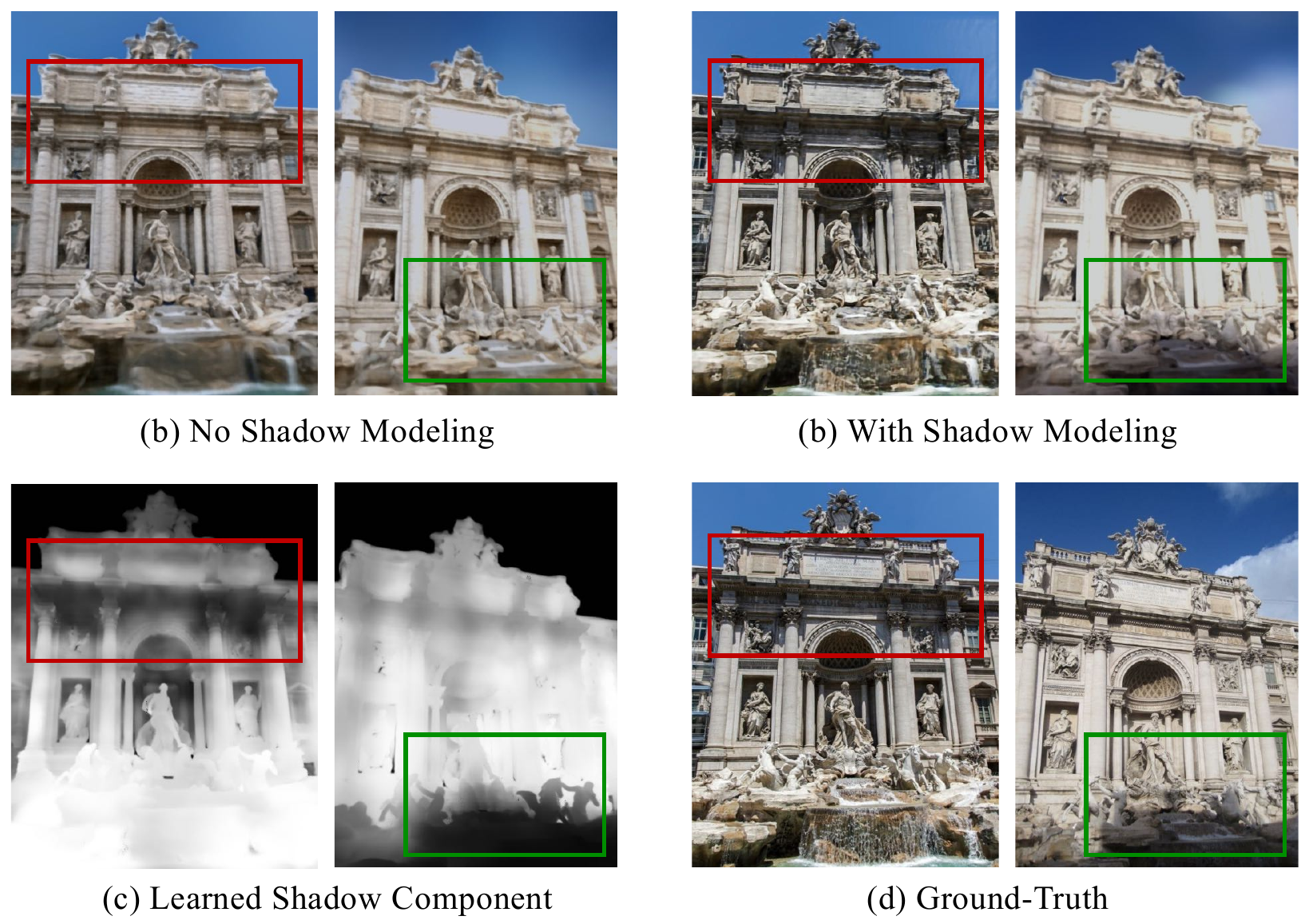}
    \vspace{-1.5em}
    \caption{We show the effectiveness of our shadow modeling and visualize the learned shadow component.
    }
    \label{fig:ablation_shadow}
    \vspace{-1em}
\end{figure}

\begin{table}[t!]
\centering
\vspace{-1.15em}
\resizebox{1.0\linewidth}{!}{
\tabcolsep 14pt
\begin{tabular}{lccc}
\toprule
\multicolumn{1}{c}{\multirow{2}{*}{Config.}} & \multicolumn{3}{c}{\textit{Trevi Fountain}} \\ \cmidrule(lr){2-4} 
\multicolumn{1}{c}{} & \multicolumn{1}{l}{PSNR $\uparrow$} & \multicolumn{1}{l}{SSIM $\uparrow$} & \multicolumn{1}{l}{LPIPS $\downarrow$}  \\ \hline
w/o Tone Mapping & 22.95 & 0.787 & 0.356 \\
w/o Shadow & 21.77 & 0.772 & 0.378 \\
w/o Realism Aug.  & 20.80 & 0.681 & 0.512 \\
Full Model & \textbf{23.09} & \textbf{0.792} & \textbf{0.345}  \\
\bottomrule
\end{tabular}
}
% \vspace{0.2em}
\caption{
We perform ablation studies of the rendering component and the realism augmentation on the \textit{Trevi Fountain}.
}
\vspace{-1.1em}
\label{tab:ablation}
\end{table}

\noindent\textbf{Composited training with occlusion-free supervision.}
We first inspect the effectiveness of the occlusion-free supervision in the re-rendering stage during composited training (Sec.~\ref{ssec:method_train}).
As shown in Fig.~\ref{fig:ablation_occlusion_free}, when training with raw images that contains occluders, we might end up with a neural model that brings some shadows at the lower part of the rendered image (Fig.~\ref{fig:ablation_occlusion_free} (a)), while the rendered result (Fig.~\ref{fig:ablation_occlusion_free} (b)) from the full model is free of such artifacts.
This proves the necessity of occlusion-free supervision for learning factorized re-rendering.

\noindent\textbf{Affine tone mapping.}
We then analyze the impact of affine tone mapping in the factorized re-rendering.
As shown in Fig.~\ref{fig:ablation} and Tab.~\ref{tab:ablation}, the rendered scene without tone mapping shows pale lighting effects compared to the ground-truth image, which demonstrates that relying on the HDR decoder alone cannot guarantee faithful modeling of various lighting effects.
By introducing affine tone mapping, we mitigate the pressure of the HDR decoder, and achieve better photo adaptation ability (\eg, Fig.~\ref{fig:ablation} (c) shows the scene lighted by yellow sunlight as the ground-truth while Fig.~\ref{fig:ablation} (a) fails).

\noindent\textbf{Shadow modeling.}
We also study the effectiveness of shadow modeling in Fig.~\ref{fig:ablation_shadow} and Tab.~\ref{tab:ablation}.
It is noticeable that when introducing shadow modeling,  our method can better simulate shadows in the outdoor scene (\eg, dimming appearance at the highlighted area in Fig.~\ref{fig:ablation_shadow}), even if the shadow caster (\eg, the shadow caused by some building behind the fountain, see highlighted green rectangle in Fig.~\ref{fig:ablation_shadow}) is not observed before.
Meanwhile, when equipped with shadow modeling, the metric of the rendering quality is also improved (see the second and the last row of Tab.~\ref{tab:ablation}).

\noindent\textbf{Realism augmentation.}
We finally inspect the efficacy of the realism augmentation mechanism.
As shown in Fig.~\ref{fig:ablation} and Fig.~\ref{fig:pipeline_adaptation}, the texture details of buildings and scenes (\eg, springs of fountain and curves of sculptures) have been enhanced with rich patterns, and the lighting effect is also much closer to the ground-truth image.
Besides, the metric result is also improved by a large margin as show in Tab.~\ref{tab:ablation}. %(see the last row of Tab.~\ref{tab:ablation}).
These results demonstrate the value of realism augmentation in broader applications such as photo adaptation and extrapolation where a single neural model is required to adapt to different illumination conditions and dynamic scene details.

\section{Conclusion}
We propose a novel factorized neural re-rendering model, which encodes the appearance and geometry of outdoor scenes from Internet photo collections in a factorized paradigm, and delivers controllable scene re-rendering, photo extrapolation and even extrapolated 3D photo generation.
One limitation is that we take the Lambertian reflectance assumption for modeling, which is not capable of representing shiny and mirrored materials such as glass walls of buildings.
Second, our method is agnostic to unobserved shadow casters (\eg, building behind the tourists), so the shadow effect is not controllable through external lighting.
Third, we do not extrapolate photos for uncaptured part of human bodies. A possible workaround is to adopt portrait image completion techniques~\cite{wu2019deepportrait} to complete the bodies, which can be directly incorporated with our pipeline in future works.

{\small\boldparagraph{Acknowledgement.} This work was partially supported by the NSFC (No.~62102356) and Zhejiang Lab (2021PE0AC01).}

% \begin{acks}
% \end{acks}

\clearpage

\bibliographystyle{ACM-Reference-Format}
\bibliography{main}

\end{document}